\documentclass[lettersize,journal]{IEEEtran}

\usepackage{amsmath,amsfonts}
\usepackage{algorithm}
\usepackage{array}
\usepackage[caption=false,font=normalsize,labelfont=sf,textfont=sf]{subfig}
\usepackage{textcomp}
\usepackage[utf8]{inputenc}
\usepackage{algpseudocode}
\usepackage{booktabs}

\usepackage{amssymb}
\usepackage{amsfonts}
\usepackage{stfloats}
\usepackage{url}
\usepackage{verbatim}
\usepackage{graphicx}
\usepackage{cite}
\usepackage{amsmath}
\usepackage{amssymb}
\usepackage{graphicx}
\usepackage{enumitem}
\usepackage{mathtools, nccmath} 
\usepackage{pifont}   
\newcommand{\xmark}{\ding{55}} 
\usepackage[utf8]{inputenc}
\usepackage{bbm}
\usepackage{multirow}
\usepackage{balance}
\usepackage{comment}
\usepackage[dvipsnames]{xcolor}

\usepackage[switch]{lineno}

\usepackage{multirow}

\begin{document}

\title{UtVAA: Ultra-tiny Vision Transformer with Affix Attention for Mobile Image Classification}

\author{Romiyal George, ~\IEEEmembership{Member,~IEEE,}
        Sathiyamohan Nishankar,  ~\IEEEmembership{Member,~IEEE,}
		Selvarajah Thuseethan, ~\IEEEmembership{Member,~IEEE,}
        Roshan G. Ragel, ~\IEEEmembership{Senior Member,~IEEE}
        
		\IEEEcompsocitemizethanks{\IEEEcompsocthanksitem Correstponding Author: S. Thuseethan (thuseethan.selvarajah@cdu.edu.au)
			\IEEEcompsocthanksitem Romiyal George, Sathiyamohan Nishankar and Roshan G. Ragel are with the University of Peradeniya, Sri Lanka. Selvarajah Thuseethan is with Charles Darwin University, Australia.}
		\thanks{Manuscript received February 15, 2020; revised XXXXXX XX, XXXX.}}

\markboth{Journal of \LaTeX\ Class Files,~Vol.~14, No.~8, August~2021}%
{Shell \MakeLowercase{\textit{et al.}}: A Sample Article Using IEEEtran.cls for IEEE Journals}


\maketitle

\begin{abstract}
Vision Transformers (ViTs) have demonstrated strong representation capability in image classification. However, their quadratic self-attention complexity and large parameter counts limit deployment on resource-constrained mobile and edge devices. This paper introduces \textbf{UtVAA}, an ultra-tiny Vision Transformer architecture designed for efficient visual recognition under strict computational budgets. It incorporates a novel \textit{Affix Attention} block that combines depthwise–pointwise local feature extraction, linear self-attention, coordinate attention for spatial dependency modelling, and a lightweight ternary fusion strategy to integrate local and global representations. In addition, Dilated Bottleneck blocks expand the receptive field using dilated depthwise separable convolutions while maintaining low FLOPs and stable optimisation through residual connections. \textbf{UtVAA} is implemented in scalable Tiny, Medium, and Large variants, with the smallest model containing 204.67K parameters and 53.95M FLOPs. Experimental results on CIFAR-10, CIFAR-100, PlantVillage-Tomato and SLIF-Tomato datasets show that \textbf{UtVAA} achieves competitive accuracy within a sub-million-parameter regime. Overall, the results demonstrate that transformer-based vision models can be redesigned into ultra-tiny architectures without significant loss in discriminative performance, making \textbf{UtVAA} suitable for mobile and edge deployment. \textcolor{blue}{Code is available at https://github.com/romiyal/UtVAA}
\end{abstract}


\section{Introduction}
\IEEEPARstart{I}{n} computer vision tasks, convolutional neural networks (CNNs) have become widely established as general-purpose backbone architectures \cite{li2021survey, cong2023review}. Since the introduction of AlexNet \cite{krizhevsky2012imagenet}, successive CNN architectures, such as VGGNet \cite{simonyan2014very}, ResNet \cite{he2016deep} and DenseNet \cite{huang2017densely}, have driven substantial advances in the field. More recently, the Vision Transformer (ViT) \cite{dosovitskiy2020image} has emerged as a compelling alternative that adapts the transformer architecture from sequence modelling to vision tasks by capturing global representations through self-attention, which explicitly models long-range dependencies across the feature map. However, despite their strong performance, traditional CNN architectures pose challenges for real-world deployment due to their large parameter counts, high floating-point operations (FLOPs) and substantial model sizes \cite{zawish2024complexity}. While ViT models often achieve state-of-the-art performance compared to CNNs, they typically require even more parameters, larger model sizes and higher FLOPs, which further limits their deployment on resource-constrained edge devices \cite{dosovitskiy2020image}. Consequently, both CNNs and ViTs demand substantial computational resources, motivating the development of more efficient vision architectures \cite{wu2023pslt}.

Recently, there has been a shift from computationally intensive CNNs and ViTs toward lightweight architectures that enable efficient deployment on resource-constrained devices \cite{liu2024lightweight}. Lightweight models have demonstrated strong potential to achieve computational efficiency while maintaining competitive performance by explicitly reducing model size and complexity through architectural design. Representative architectures such as MobileNet \cite{howard2017mobilenets}, ShuffleNet \cite{zhang2018shufflenet} and Mobile-ViT \cite{mehta2021mobilevit} facilitate the practical deployment of deep learning solutions in resource-limited settings. Nevertheless, existing lightweight models often struggle with effective multi-scale feature extraction, achieving consistently high accuracy, and maintaining adaptability across diverse and complex scenarios \cite{olimov2023consecutive, addo2024hybrid, george2025background, lu2025general}. Moreover, the rapid expansion of industrial applications further intensifies the need to continuously reduce computational overhead, such as parameter counts, FLOPs and overall operational costs \cite{liang2021cemodule, xiang2021novel}.

Non-architectural model compression methods, such as pruning, quantisation and knowledge distillation, aim to reduce model size and computational overhead to facilitate deployment on resource-constrained devices, but often incur degradation in feature representation capacity and predictive performance \cite{ji2022neural, liu2024lightweight}. In contrast, lightweight architectural designs, although inherently constrained in representational capacity, offer modular structures that enable the seamless integration of external feature-enhancement mechanisms. Attention mechanisms have consistently demonstrated effectiveness in strengthening feature representation and improving classification accuracy across computer vision tasks \cite{guo2022attention}. Accordingly, recent CNN and ViT based lightweight models increasingly incorporate attention modules, including Coordinate Attention (CA) \cite{hou2021coordinate} as adopted in \cite{sun2025lightweight}, the Convolutional Block Attention Module (CBAM) \cite{woo2018cbam} in \cite{guo2025lwheatnet, ferdous2025mednet}, Squeeze-and-Excitation (SE) \cite{hu2018squeeze} in \cite{dash2024seimb} and Efficient Channel Attention (ECA) \cite{wang2020eca} in \cite{duan2025improving}. As modern neural architectures increasingly prioritise parameter efficiency, they become well-suited to resource-constrained deployment. Although the integration of attention modules modestly increases the parameter count, deviating from a strictly lightweight paradigm, it yields substantial improvements in feature extraction and predictive accuracy. Recent linear attention variants further alleviate this trade-off by preserving computational efficiency in lightweight networks.

Leveraging recent advances in lightweight architectures, there is an emerging shift toward ultra-compact networks, exemplified by models such as SqueezeNet \cite{iandola2016squeezenet}, which drastically reduce parameter counts while maintaining competitive accuracy. Neural network architectures such as MobileNet variants and ShuffleNet variants span the spectrum of lightweight models, typically comprising one million or more parameters. In contrast, the ultra-compact network SqueezeNet contains less than one million parameters, achieving an order-of-magnitude reduction compared to many lightweight designs while maintaining competitive performance. Transformer-based vision networks, despite consistently outperforming CNNs, remain largely unexplored in this ultra-lightweight regime. Despite their consistent superiority over CNNs, transformer-based vision networks have received limited attention in ultra-lightweight architectures. MobileViT partially addresses this gap by integrating lightweight convolutional layers with transformer blocks to capture both local and global representations in a parameter-efficient manner \cite{mehta2021mobilevit}. To date, there has been no systematic investigation into the design of tiny transformer models that maintain strong feature representation and predictive performance while minimising both parameter count and computational cost. This motivates the development of ultra-tiny ViT architectures that maintain high accuracy with practical deployment on resource-limited devices.

\begin{figure*}
  \begin{center}
  \includegraphics[width=\textwidth]{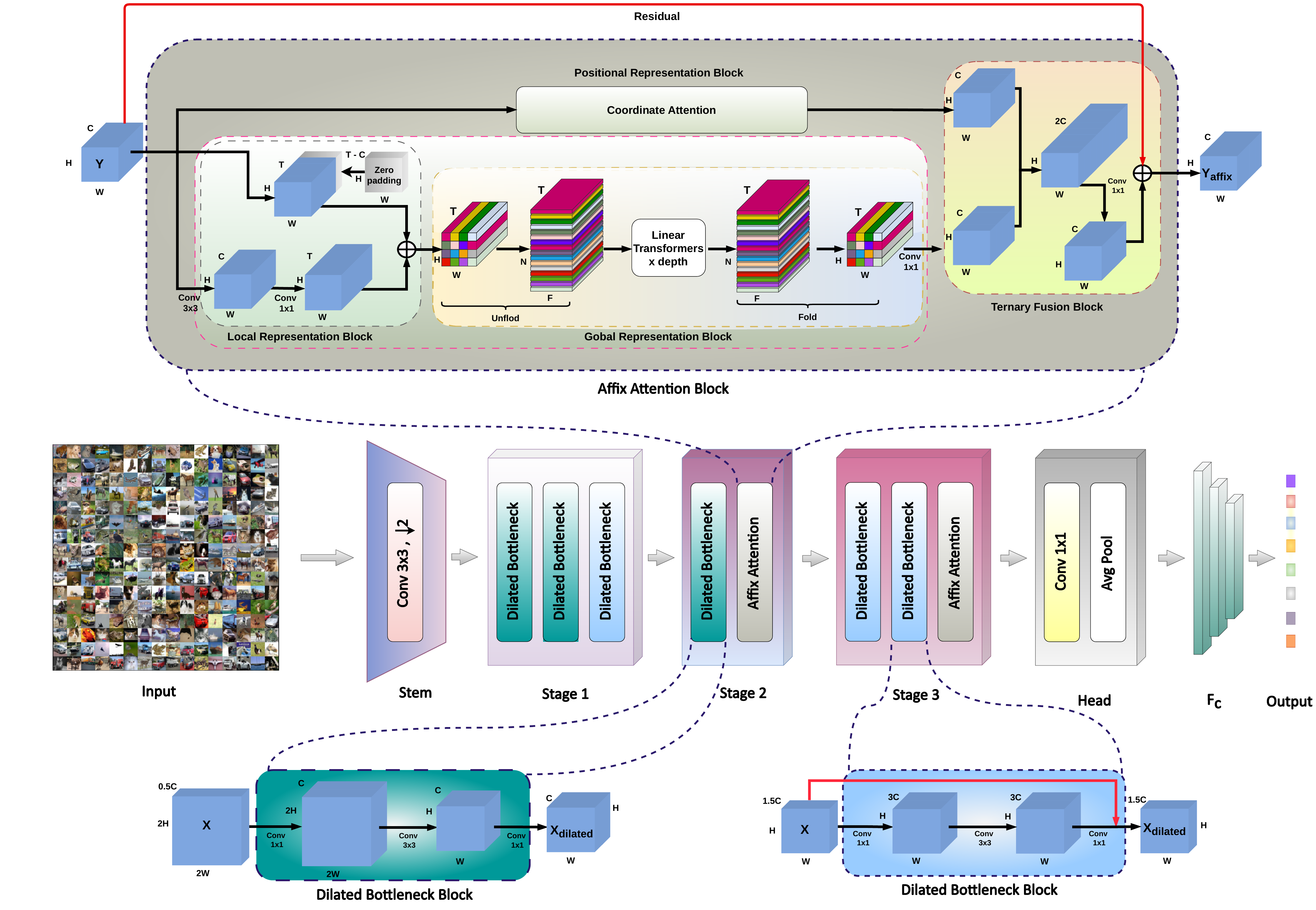}
  \caption{The overall architecture of the proposed ultra-tiny UtVAA framework for mobile image classification.}
  \label{fig:overallarchitecture}
  \end{center}
\end{figure*}

To address the increasing demand for image classification models that balance strong feature extraction capacity with limited resource constraints, this paper proposes \textbf{UtVAA}, an \textbf{U}ltra-\textbf{t}iny \textbf{V}iT architecture that incorporates a novel \textbf{A}ffix \textbf{A}ttention mechanism for efficient mobile image classification, as shown in Fig. \ref{fig:overallarchitecture}. A key limitation of existing ViT designs arises from the direct integration of conventional self-attention modules, which becomes computationally prohibitive under aggressive model scaling. In ultra-tiny architectures, the quadratic computational cost and high memory footprint of standard self-attention operations significantly hinder scalability and inference efficiency on resource-limited devices \cite{bravin2025embbert, fan2025ultralbm}. This inefficiency often negates the gains achieved from parameter reduction and limits the practical deployment of ViT models in mobile and edge environments. In response, UtVAA introduces the novel Affix Attention mechanism and a Dilated Bottleneck to enhance salient feature interactions while incurring negligible parameter overhead in a selective manner. By affixing lightweight attention cues to intermediate visual feature representations, the proposed design preserves global contextual modelling while maintaining ultra-tiny model complexity suitable for edge deployment. In contrast to non-architectural compression strategies that often compromise feature expressiveness, UtVAA adopts a principled architectural design to balance computational efficiency and classification accuracy. This study demonstrates that ViT models can be systematically redesigned as ultra-tiny architectures for high-performance mobile image classification. The major contributions of this work are summarised below.

\begin{enumerate}
    \item Proposes a novel attention block, namely the \textbf{Affix Attention Block}, which jointly encodes global and local positional information with residual feature representations using an efficient linear self-attention formulation. A coordinate attention mechanism is incorporated to capture fine-grained spatial dependencies, while residual connections ensure stable feature propagation.

    \item Proposes two novel \textbf{Dilated Bottleneck Blocks}, with and without residual connections. Both designs employ dilated depthwise and pointwise convolutions to effectively expand the receptive field at low computational cost, while the residual variant further improves optimisation stability and classification accuracy.
    
    \item Proposes a novel ultra-tiny architecture, namely \textbf{UtVAA}, which systematically integrates the proposed Affix Attention Blocks and Dilated Bottleneck Blocks, as illustrated in Fig. \ref{fig:overallarchitecture}. UtVAA achieves low FLOPs and parameter counts while retaining competitive accuracy for resource-constrained image classification.

    \item UtVAA achieves competitive performance on two standard image classification benchmarks, \texttt{CIFAR-10} and \texttt{CIFAR-100}, as well as on two widely used plant disease classification datasets, namely \texttt{PlantVillage} \texttt{SLIF-Tomato}.
\end{enumerate}

The remainder of this paper is organized as follows. Section \ref{sec:relatedwork} reviews representative lightweight architectures and attention mechanisms for image classification. Section \ref{sec:proposedmethod} presents the proposed UtVAA in detail. Section \ref{sec:Experiments} reports the experimental results, complexity analysis, and discussion. Finally, Section \ref{sec:conclusion} concludes the paper and outlines future research directions.

\section{Related work}\label{sec:relatedwork}
This section provides an overview of lightweight architectures for image classification on resource-constrained devices. In addition, it reviews attention mechanisms that further enhance the performance of these models.

\subsection{Lightweight Neural Networks}
A decade ago, CNNs became the standard for image classification following the success of AlexNet \cite{krizhevsky2012imagenet}. Deeper network architectures were later introduced to enhance feature representation, including VGG \cite{simonyan2014very}, which showed that stacking small convolutional filters (i.e., $3 \times 3$) can be effective. ResNet \cite{he2016deep} used residual blocks and skip connections to mitigate the degradation issue in deep networks. DenseNet \cite{huang2017densely} established connections from each layer to all following layers to maximize information propagation. The primary goal of these models is to refine feature extraction and achieve higher classification accuracy. Despite their accuracy, large networks with standard convolutional layers are often impractical for edge devices due to high memory demands \cite{george2025past, zhang2025breaking, ngo2025edge}. In parallel with CNN developments, the transformer architecture was successfully applied to vision tasks, as ViT \cite{dosovitskiy2020image}. This line of work led to models such as PiT \cite{heo2021rethinking}, DeiT \cite{touvron2021training} and Swin Transformer \cite{liu2021swin}. The high computational demand of many ViT architectures, as observed for CNNs, constrains their deployment on resource-limited platforms. Such challenges prompted increased attention to deep neural network designs suitable for resource-limited environments.

Lightweight architectures facilitate model deployment on resource-limited hardware. The use of depthwise and pointwise convolutions in place of standard convolutions led to lightweight architectures such as MobileNet \cite{howard2017mobilenets}. In SqueezeNet \cite{iandola2016squeezenet}, the Fire module reduces parameters by compressing channels with a squeeze layer followed by expansion using $1 \times 1$ and $3 \times 3$ convolutions. ShuffleNet \cite{ma2018shufflenet} reduces computation by combining group convolutions, channel shuffling, and depthwise convolutions in its bottleneck layers. Neural Architecture Search (NAS) automates network design and has been shown to yield high-performance models. EfficientNet \cite{tan2019efficientnet} leverages NAS to scale network depth, width, and resolution together. In practice, NAS requires substantial computational resources and offers limited generalization \cite{salmani2025systematic}. Compression of pre-trained models is also widely used to reduce computational cost. Model compression approaches include pruning, quantization, and knowledge distillation, which transfers learned behavior from a large model to a smaller one.

Recent studies in lightweight neural networks move toward sub‑million‑parameter \texttt{tiny} networks that can operate within the memory constraints of micro controllers. Existing surveys report that many lightweight CNNs still exceed one million parameters, which restricts their use on ultra-low-power micro-controllers and drives the need for smaller architectures \cite{somvanshi2025tiny}. Results from tiny CNN designs suggest that strong performance on MNIST-scale tasks can be achieved with only a small memory usage \cite{isong2025building}. For more complex vision tasks, PiDiNet shows that pixel-difference and binary convolutions can attain good accuracy with $\approx 1$ million parameters, improving efficiency beyond standard depthwise separable designs \cite{su2023lightweight}. Recent Tiny deep learning NAS frameworks and differentiable NAS approaches consider multiple constraints, such as accuracy, energy use, latency, and memory, to produce ImageNet-scale models that fit within limited on-chip memory \cite{burrello2023enhancing}. Nonetheless, existing literature points to accuracy loss, weak correlation between FLOPs counts, and limited cross-platform generalization as key limitations of tiny architectures \cite{liu2024lightweight}. These limitations have driven interest in obtaining specialised tiny networks that, such as SqueezeNet, stay below one million parameters while remaining accurate under severe resource constraints.

\begin{table*}[h]
\caption{Detailed architecture of UtVAA with comparison across Tiny, Medium, and Large versions.}
\label{tab:modeldetails}
\centering
\renewcommand{\arraystretch}{1.3}
\begin{tabular}{llccccc}
\hline
\multirow{2}{*}{\textbf{Stage}} & \multirow{2}{*}{\textbf{Layer}} & \multirow{2}{*}{\textbf{Output Size}} & \multirow{2}{*}{\textbf{Output Stride}} & \multicolumn{3}{c}{\textbf{Output Channels}} \\ \cline{5-7}
 &  &  &  & \textbf{Tiny} & \textbf{Medium} & \textbf{Large} \\
\hline
Input & Image & 256 × 256 & 1 & - & - & - \\
\hline
Stem & Conv 3 × 3, ↓2 & 128 × 128 & 2 & 8 & 8 & 8 \\
\hline
\multirow{3}{*}{Stage 1} & Dilated Bottleneck, ↓2 & 64 × 64 & 4 & 8 & 8 & 8 \\
 & Dilated Bottleneck, ↓2 & 32 × 32 & 8 & 16 & 32 & 32 \\
 & Dilated Bottleneck & 32 × 32 & 8 & 16 & 32 & 32 \\
\hline
\multirow{2}{*}{Stage 2} & Dilated Bottleneck, ↓2 & 16 × 16 & 16 & 32 & 64 & 72 \\
 & Affix Attention Block & 16 × 16 & 16 & 32 (d = 48) & 64 (d = 96) & 72 (d = 112) \\
\hline
\multirow{3}{*}{Stage 3} & Dilated Bottleneck & 16 × 16 & 16 & 48 & 96 & 128 \\
 & Dilated Bottleneck & 16 × 16 & 16 & 48 & 96 & 128 \\
 & Affix Attention Block & 16 × 16 & 16 & 48 (d = 64) & 96 (d = 112) & 128 (d = 132) \\
\hline
\multirow{2}{*}{Head} & Conv 1 × 1 & 16 × 16 & 16 & 288 & 352 & 416 \\
 & Global Pool & 1 × 1 & 256 & 288 & 352 & 416 \\
\hline
Classifier & Linear & 1 × 1 & 256 & 100 & 100 & 100 \\
\hline \hline
\multirow{2}{*}{Complexity} & Parameters (K)  &  &  & 204.67 & 609.36 & 858.13 \\ 
 & FLOPs (M) &  &  & 53.95 & 159.24 & 220.94 \\
\hline
\end{tabular}
\end{table*}

\subsection{Attention Mechanisms}
Attention mechanisms are often applied to improve performance in lightweight neural networks \cite{hu2024hybrid}. For instance, the Context-Aware Decoupled Fully Connected (CADFC) attention mechanism enhances MobileNet by adding a context-aware dual-attention structure \cite{xue2025lightweight}. Yan et al. \cite{yan2024convolutional} reported 98.74\% accuracy in cancer classification using a lightweight model with carefully integrated attention mechanisms. Bendelhoum et al. \cite{bendelhoum2025enhancing} integrated coordinate attention into MobileNetV3 and reported an accuracy of 99.28\% for facial expression recognition. Using soft attention and depthwise separable convolutions, a faster and lightweight CNN model achieved 99.30\% accuracy on the PlantVillage tomato leaf disease dataset \cite{alnamoly2024fl}. Efficient Local Attention (ELA) encodes positional information using one-dimensional convolutions without channel reduction and improves ImageNet performance when applied to the MobileNet architecture. Sequential Fusion Attention (SFA) is a lightweight approach that improves ImageNet accuracy by combining efficient spatial and channel attention within strict computational limits \cite{xu2025sfa}.

Lightweight attention-based networks are widely used in imaging and sensing tasks to improve discrimination while operating under strict model size and computational constraints. Recent work increasingly co‑optimises attention design and network hierarchy for domain‑specific lightweight classification. LightFormer employs a shallow hybrid attention block followed by a lightweight stage, achieving over 95\% accuracy on remote sensing tasks with low memory and computational requirements \cite{wang2025lightformer}. Lightweight CNN and Deeplab-style models with attention are widely used in remote sensing tasks where computational resources are limited \cite{tong2020channel}. Lightweight spectral–spatial attention, often paired with 3D convolutions, is widely used in hyperspectral and few-shot classification when labelled samples are scarce \cite{guo2025edb}. Recent work on lightweight attention-based pest recognition, which extends MobileNetV2 with a simplified residual attention mask to achieve over 96\% accuracy on multiple field pest datasets while remaining deployable on smartphones \cite{janarthan2024liran}. In a similar work, a complementary double attention-based lightweight network that introduces dual attention modules to better distinguish visually similar pests and consistently outperforms existing lightweight models on both small and large benchmark datasets \cite{janarthan2025efficient}.

On the other hand, lightweight ViT designs focus on efficient attention and token mixing to make image classification feasible on resource-limited devices. Convolutional Additive Self-attention ViT (CAS-ViT) introduces an additive token mixer in place of quadratic self-attention to support efficient mobile-oriented vision backbones \cite{zhang2024cas}. In another work, sparse attention is paired with depthwise separable convolutions and a channel-interactive feed-forward module to improve efficiency without losing multi-scale context \cite{zhang2025lightweight}. To further reduce complexity, some methods restrict attention across space or depth. Pyramid Window-based Lightweight Transformer (PWLT) introduces pyramid window-based local–global attention with dual self-attention to capture cross-window relationships \cite{mo2024pwlt}. In contrast, depth-wise convolutional shortcuts introduce local inductive bias to ViTs, improving convergence, especially for limited data \cite{zhang2025depth}. Some domain-specific variants embed lightweight attention into small ViT architectures, such as MobileViT and DeiT-Tiny, and further employ fuzzy attention to enhance class-relevant token selection when computational resources are constrained in medical imaging \cite{chandra2025fuzzy}.

\section{Proposed Method}\label{sec:proposedmethod}
The proposed \textbf{UtVAA} architecture, illustrated in Fig. \ref{fig:overallarchitecture} and Table \ref{tab:modeldetails}, adopts a hybrid design that initially captures fine-grained local representations and subsequently integrates global contextual modelling with positional information to characterise long-range dependencies while maintaining computational efficiency effectively. \textbf{UtVAA} adopts a hierarchical multi-stage architecture inspired by ViT and CNN designs, which facilitates progressive spatial downsampling and the extraction of increasingly high-level semantic features. Furthermore, residual connections are systematically incorporated to stabilise optimisation and enhance the discriminative capacity of block-level feature representations. Table \ref{tab:modeldetails} summarises the \textbf{UtVAA} variants of different sizes, such as \texttt{Tiny}, \texttt{Medium}, and \texttt{Large}, which differ in the number of output channels in the intermediate layers. These configurations comprise 204.67K, 609.36K, and 858.13K trainable parameters, respectively.

\subsection{Architectural Details}
This subsection outlines the design details of the proposed stages of the \textbf{UtVAA} framework.

\subsubsection{Stem}
The stem applies a $3 \times 3$ convolutional layer with a stride of 2 to perform downsampling. By reducing the spatial resolution early in the network, the model decreases computational demand \cite{hesse2023content} while preserving key low-level feature representations \cite{bazgir2020representation}. The output feature map $X_\text{stem} \in \mathbb{R}^{B \times C_{o} \times H/2 \times W/2}$ can be expressed as:

\begin{equation}
X_\text{stem} = \phi\Big(\mathcal{B}(\mathcal{C}_{3\times 3, 2}^{3 \to C_{o}}(I))\Big)
\end{equation}
where $I \in \mathbb{R}^{B \times 3 \times H \times W}$ denotes the input batch with batch size $B$ and spatial dimensions $H \times W$. The operator $\mathcal{C}_{3\times3, 2}^{3 \to C_{o}} (\cdot)$ represents a $3 \times 3$ convolution with stride 2, transforming the input from 3 channels to $C_{o}$ output channels. In all variants of the \texttt{UtVAA} framework, $C_{o} = 8$. The functions $\mathcal{B}$ and $\phi$ correspond to Batch Normalisation and the SiLU activation, respectively.

Using a stride of 2 reduces the spatial dimensions from $H \times W$ to $\frac{H}{2} \times \frac{W}{2}$. The computational complexity of the $3 \times 3$ convolution in the stem stage is given by $3^2 \cdot \frac{H}{2} \cdot \frac{W}{2} \cdot 3 \cdot 8$, accounting for the reduced spatial resolution due to downsampling. This results in an approximate $4\times$ reduction in FLOPs, along with a corresponding $4\times$ reduction in the memory footprint of feature maps. Consequently, the convolution layer in the stem significantly reduces the computational burden of subsequent network stages while preserving essential feature information. 

\begin{algorithm}
\caption{Dilated Bottleneck}\label{alg:dilatedbottleneck}
\begin{algorithmic}[1]
\Require $X \in \mathbb{R}^{B \times C_i \times H \times W}$, $C_i$, $C_m$, $C_o$, $s$, $d$
\Ensure $X_{dilated} \in \mathbb{R}^{B \times C_o \times H' \times W'}$
\State \textbf{Padding Selection:}
\If{$s = 1$}
    \State \hspace*{1em} $p \gets d$
\ElsIf{$s = 2$}
    \State \hspace*{1em} $p \gets \begin{cases} 
        1 & \text{if } d = 1 \\
        2 & \text{if } d = 2 \\
        \lfloor (2d - 1)/2 \rfloor & \text{otherwise}
    \end{cases}$
\EndIf
\State \textbf{Residual Setup:}
\State \hspace*{1em} $r \gets (C_i = C_o \land s = 1)$
\State \hspace*{1em} $X_{res} \gets X$ 
\State \textbf{Bottleneck:}
\State \hspace*{1em} $X_1 \gets \text{PointwiseConv}(C_i, C_m, 1 \times 1)(X)$
\State \hspace*{1em} $X_2 \gets \text{DepthwiseConv}(C_m, C_m, 3 \times 3, s, p, d)(X_1)$
\State \hspace*{1em} $X_3 \gets \text{PointwiseConv}(C_m, C_o, 1 \times 1)(X_2)$
\State \textbf{Residual Connection:}
\If{$r \land \text{shape}(X_{res}) = \text{shape}(X_3)$}
    \State \hspace*{1em} $X_{dilated} \gets X_3 + X_{res}$
\Else
    \State \hspace*{1em} $X_{dilated} \gets X_3$
\EndIf
\State \Return $X_{dilated}$
    
{\footnotesize
\Statex \rule{\linewidth}{0.4pt}
$B$ - batch size; 
$H \times W$ - Input spatial dimensions; 
$H' \times W'$ - Output spatial dimensions; 
$C_i$ - Input channels; 
$C_m$ - Middle/expanded channels; 
$C_o$ - Output channels; 
$s$ - Stride; 
$d$ - Dilation rate; 
$p$ - Padding. 
}
\end{algorithmic}
\end{algorithm}

\subsubsection{Dilated Bottleneck Block}
Two variants of the Dilated Bottleneck block are proposed for efficient local feature extraction with low computational cost: one with a residual connection and the other without. Their modular design allows flexible integration across different network configurations. The block combines pointwise and dilated depthwise convolutions with residual connections, which reduces computational complexity relative to standard convolutions \cite{sandler2018mobilenetv2, lei2019dilated}.

Algorithm \ref{alg:dilatedbottleneck} details the Dilated Bottleneck block. At line 11, the process begins with a $1 \times 1$ pointwise convolution to expand the channel dimension, followed by a $3 \times 3$ depthwise convolution with configurable stride and dilation. A final $1 \times 1$ pointwise convolution then projects the channels to the target dimension. This block is applied across multiple stages, with spatial resolution reduced at stage transitions via downsampling. The use of dilation enlarges the receptive field without increasing the number of parameters, which is particularly advantageous in deeper layers \cite{yu2015multi, george2025background}. When the input and output dimensions are matched, a residual connection is incorporated to facilitate gradient flow that mitigates the vanishing-gradient problem \cite{he2016deep} and supports the training of deeper architectures. The output feature map from the Dilated Bottleneck block $X_{dilated}$ is calculated using the equation below.

\begin{equation}
X_{dilated} = \delta X_{res} + \mathcal{P}_{1 \times 1}^{C_m \to C_o}
\Big(\mathcal{D}_{3 \times 3}^{C_m \to C_m}
(\mathcal{P}_{1 \times 1}^{C_i \to C_m}(X))\Big)
\end{equation}
where, $\mathcal{P}_{1 \times 1} ^ {C_i \to C_m} (\cdot)$ and $\mathcal{D}_{3 \times 3} ^ {C_m \to C_m} (\cdot)$ denotes pointwise convolution and depthwise convolution, respectfully. The $\delta \in \{0,1\}$ indicates the absence or presence of the residual connection. The residual connection is applied only when $C_\text{i}=C_\text{o}$ and $s = 1$.

The proposed dilated bottleneck achieves a computational complexity of approximately $H \cdot W \cdot C_m \left(C_i + \frac{9 + C_o}{s^2}\right)$, which is substantially lower than that of stacking three conventional $3 \times 3$ convolutions. In comparison, the proposed design reduces the computational cost by approximately $\frac{C_o}{C_m} \cdot \frac{s^2}{3}$ times, while preserving a comparable receptive field through dilation. Consequently, the Dilated Bottleneck reduces computational complexity by approximately 8–9 times relative to standard convolutions, while dilation expands the receptive field without additional overhead. This design facilitates the capture of multi-scale contextual information, which is critical for accurate recognition of diverse patterns in image classification.

\subsubsection{Affix Attention Block}
The principal novel component of \textbf{UtVAA} is the Affix Attention Block, which is inspired by the real-world principle of \textit{Synergistic Affixion} and adopted as the conceptual foundation for its design. The Affix Attention Block mitigates the loss of spatial information commonly associated with linear self-attention. In contrast to sequential lightweight hybrid designs, such as MobileViT \cite{mehta2021mobilevit}, it adopts a parallel architecture that jointly processes a global representation block based on a linear transformer and a positional representation block incorporating coordinate attention \cite{hou2021coordinate}. This parallel formulation preserves directional spatial dependencies while integrating global contextual information through coordinate-aware representations.

\begin{algorithm}
\caption{Affix Attention Block}
\label{alg:affixattentionblock}
\begin{algorithmic}[1]
\Require $Y \in \mathbb{R}^{B \times C_i \times H \times W}$, $C_i$, $T$, $C_o$, $F$
\Ensure $Y_{affix} \in \mathbb{R}^{B \times C_o \times H \times W}$

\State \textbf{Residual Setup:}
\State $Y_{res} \gets Y$

\State \textbf{Local Representation:}
\State $Y_1 \gets \text{DepthwiseConv}(C_i, C_i, 3 \times 3, 1)(Y)$ 
\State $Y_2 \gets \text{PointwiseConv}(C_i, T, 1 \times 1)(Y_1)$
\State $Z_{pad} \gets \text{Zeros}(B, T - C_i, H, W)$ 
\State $Y_3 \gets [Y, Z_{pad}, \text{dim}=1]$ 
\State $Y_{loc} \gets Y_2 + Y_3$

\State \textbf{Global Representation:}
\State $Y_5 \gets \text{Unfold}(Y_{loc}, (B, F^2, \tfrac{H}{F} \times \tfrac{W}{F}, T))$
\State $Y_{linear} \gets \text{LinearTransformer}(Y_5)$ 
\State $Y_6 \gets \text{Fold}(Y_{linear}, (B, T, H, W))$ 
\State $Y_{glo} \gets \text{Conv2d}(T, C_i, 1 \times 1)(Y_6)$

\State \textbf{Coordinate Attention:}
\State $Y_{coor} \gets \text{CoordinateAttention}(C_i, C_i)(Y)$

\State \textbf{Ternary Fusion:}
\State $Y_7 \gets [Y_{coor}, Y_{glo}]$ 
\State $Y_{ter} \gets \text{Conv2d}(2 \cdot C_i, C_o, 1 \times 1)(Y_7)$ 
\State $Y_{affix} \gets Y_{ter} + Y_{res}$ 
\State \Return $Y_{affix}$

{\footnotesize
\Statex \rule{\linewidth}{0.4pt}
\Statex \textbf{Notation:} 
$B$ - Batch size; 
$H \times W$ - Spatial dimensions; 
$C_i$ - Input channels;
$c_o$ - Output channels; 
$T$ - Transformer dimension; 
$F$ - Patch size; 
}
\end{algorithmic}
\end{algorithm}

It consists of a \textit{local representation block} for capturing fine-grained features, a \textit{global representation block} for modelling long-range dependencies, a \textit{positional representation block} incorporating coordinate attention to encode spatial context and a \textit{residual connection} to support stable feature propagation. A detailed description is provided in Algorithm \ref{alg:affixattentionblock}.

\begin{enumerate}[label=\Roman*.]
    \item \textbf{Local Representation Block:}
    In this block, a local feature map is generated from the input using a $3 \times 3$ depthwise separable convolution, followed by a $1 \times 1$ pointwise convolution. In both cases, a padding of 1 is applied. This operation projects the input channels into a higher-dimensional embedding space suitable for linear self-attention. The resulting feature map is concatenated with a corresponding feature map after zero-padding to ensure spatial alignment. The output feature map $Y_{loc}$ is given by: 
    \begin{equation}
        Y_{loc} = \mathcal{P}_{1 \times 1}^{C_i \to T} \Big(\mathcal{D}_{3 \times 3}^{C_i \to C_i}(Y)\Big) + [Y, \mathcal{Z}_{H \times W} ^ {T-C_i}]
    \end{equation}
    where, $\mathcal{Z}$ is zero-padding function. $Y$ is the output feature map from one of the dilated bottleneck layers $X_{dilated}$.
    
    The computational complexity of this block is dominated by the convolutional operations applied to the input feature map of size $H \times W$ with $C_i$ channels. In particular, the $3 \times 3$ depthwise convolution contributes a complexity of $H \cdot W \cdot C_i \cdot 3^2$, while the subsequent $1 \times 1$ pointwise convolution adds $H \cdot W \cdot C_i \cdot T$. This results in a total complexity proportional to $H \cdot W \cdot C_i \cdot (9 + T)$. Zero-padding and concatenation introduce negligible computational overhead relative to convolutional operations and are therefore typically excluded from complexity analysis.

    \item \textbf{Global Representation Block:}
    To support efficient modelling of long-range dependencies with minimal computational overhead, this block integrates a \textit{linear transformer} designed for mobile-friendly architectures. The input feature map is first unfolded, partitioned into non-overlapping patches, and subsequently processed by a linear transformer module. In the linear self-attention mechanism of the linear transformer, projections of the query (Q), key (K) and value (V) tensors are computed, as shown below.
    \begin{equation}
        [Q, K, V] = \text{Conv}_{1\times1}^{\text{QKV}}(Y_{5})
    \end{equation}
        
    A softmax function is applied to the Q branch to obtain context scores. These scores are then multiplied element-wise with the $K$ projections and aggregated to form a global context vector $g$, as given in the equation below.

    \begin{equation}
        g = \sum_{n=1}^{N} K \odot \text{Softmax}(Q)
    \end{equation}
    where $N$ represents the total number of patches.

    The context vector ($g$) is then broadcast across the $N$ dimensions to form $g^{\text{expand}}$. The output of the linear self-attention block is obtained by modulating the value projections with the expanded global context vector, followed by a point-wise non-linearity and a final $1 \times 1$ convolution that projects the aggregated representations back to the feature space as $Y_{\text{LSA}}$.

    \begin{equation}
        Y_{\text{LSA}} = \text{Conv}_{1\times1}\Big(\text{ReLU}(V) \odot g^{\text{expand}}\Big)
    \end{equation}

    $Y_{\text{LSA}}$ is then combined with the original input feature map $Y_{5}$ via a residual connection to obtain the attended feature $Y_{\text{attn}}$. A feed-forward network is subsequently applied to $Y_{\text{attn}}$, and its output is added through a second residual connection to produce the final representation $Y_{\text{linear}}$, enhancing feature expressiveness while maintaining stable gradient flow:
    
    \begin{align}
        Y_\text{attn} &= Y_{5} + Y_{\text{LSA}} \\
        Y_\text{linear} &= Y_\text{attn} + \text{feedfoward}(Y_\text{attn})
    \end{align}

    The linear Transformer block is applied iteratively according to a predefined depth, with repeated attention refinement and feed-forward operations. As illustrated in Fig. \ref{fig:overallarchitecture}, the first Affix Attention module is configured with depth $=2$, while the second Affix Attention block is set to depth $=3$

    $Y_{\text{linear}}$ is reshaped into a 2D spatial grid $Y_{6}$ via a folding operation. A $1 \times 1$ convolution is then applied to project the features back to the original input channel dimension, as shown below.
    \begin{equation}
        \mathbf{Y}_{glo} = \text{Conv}_{1\times1} \left( Y_{6} \right)
    \end{equation}

    The linear transformer reduces computational cost compared to standard self-attention by avoiding pairwise token interactions and using linear projections with element-wise operations. This leads to linear scaling with the number of patches.

    \item \textbf{Coordinate Attention:} The input feature map $Y \in \mathbb{R}^{B \times C_{\text{i}} \times H \times W}$ is processed through a coordinate attention mechanism that captures long-range dependencies along spatial dimensions by performing separate global pooling along the height and width axes. By incorporating positional information into channel attention, coordinate attention supports efficient modelling of spatial relationships and improves feature representation in lightweight architectures \cite{hou2021coordinate}.

    \begin{equation}
        \mathbf{M}_h = \text{AvgPool}_H(Y), \quad
        \mathbf{M}_w = \text{AvgPool}_W(Y)
    \end{equation}
    Here, $\mathbf{M}_h \in \mathbb{R}^{B \times C_{\text{in}} \times H \times 1}$ aggregates information along the width dimension to capture vertical (height-wise) context, while $\mathbf{M}_w \in \mathbb{R}^{B \times C_{\text{i}} \times 1 \times W}$ aggregates along the height dimension to capture horizontal (width-wise) context.
    
    The intermediate representation $Z$ is defined as:
    \begin{equation}
        \mathbf{Z} = \mathcal{B}\left(\mathcal{C}_{1\times1}^{C_{\text{i}} \to C_{\text{m}}}\left([\mathbf{M}_h, \text{Permute}(\mathbf{M}_w)]\right)\right)
    \end{equation}
    where $\mathcal{B}(\cdot)$ denotes batch normalisation, which stabilises the feature distribution and improves training convergence. The operator $\mathcal{C}_{1\times1}^{C_{\text{i}} \to C_{\text{m}}}(\cdot)$ represents a point-wise convolution that reduces the channel dimension from $C_{\text{i}}$ to $C_{\text{m}}$. The term $[\mathbf{M}_h, \text{Permute}(\mathbf{M}_w)]$ denotes denotes the concatenation of height-wise and width-wise pooled features

    Then, a non-linear activation (i.e., hard-swish) is applied to $\mathbf{Z}$ to enhance feature expressiveness while maintaining computational efficiency.
    
    \begin{equation}
        \mathbf{Z} = h_{\mathrm{swish}}(\mathbf{Z})
    \end{equation}

    The fused representation is partitioned into height- and width-specific feature maps to facilitate directional attention modelling.
    \begin{equation}
        \mathbf{Z}_h, \mathbf{Z}_w = \text{Split}(\mathbf{Z}, [H, W], \text{dim}=2)
    \end{equation}

    Each directional feature is processed with a $1\times1$ convolution and a sigmoid activation to produce attention weights that encode spatial importance along the height and width dimensions.

    \begin{equation}
        A_h = \sigma\!\left(\mathcal{C}_{1\times1}^{C_{\text{m}} \to C_{\text{i}}}(\mathbf{Z}_h)\right), \quad
        A_w = \sigma\!\left(\mathcal{C}_{1\times1}^{C_{\text{m}} \to C_{\text{i}}}(\mathbf{Z}_w)\right)
    \end{equation}
    where, $\sigma(\cdot)$ denotes the sigmoid activation function, which maps the output of the $1\times1$ convolution to the range $(0,1)$.

    Finally, the attention maps are applied to the input feature map via element-wise multiplication, adaptively reweighting features according to spatial relevance.
    \begin{equation}
        \mathbf{Y}_{\text{coor}} = \mathbf{Y} \odot A_h \odot A_w
    \end{equation}
    
    The computational complexity of Coordinate Attention is given by $\mathcal{O}(H \cdot W \cdot C_{\text{i}} + (H+W)\cdot \frac{C_{\text{i}}^2}{r})$, where the first term corresponds to spatial feature aggregation across height and width, and the second term arises from the two $1\times1$ convolutional bottleneck operations with channel reduction ratio $r$.

    \item \textbf{Ternary Fusion Block :} 
    The Ternary Fusion Block is a key sub-module of the Affix Attention Block, which concatenates feature maps from the Global Representation Block, Coordinate Attention and the residual connection.

    \begin{equation}
        Y_{\text{affix}} = \mathcal{C}_{1\times1}^{2C_{\text{i}} \rightarrow C_{\text{o}}}([Y_{\text{glo}},\, X_{\text{coor}}]) + Y
    \end{equation}

\end{enumerate}

The proposed \textbf{UtVAA} framework combines lightweight convolutions with structured attention to efficiently capture local, global and directional spatial features. It integrates dilated bottlenecks, affix attention, and coordinate attention to achieve multi-scale representation while maintaining stable training and low computational cost. This balance of accuracy and efficiency makes the model suitable for resource-constrained vision tasks.

\section{Experiments}\label{sec:Experiments}
This section evaluates the proposed \textbf{UtVAA} architecture in terms of classification accuracy, computational efficiency and suitability for deployment on resource-constrained devices. 

\subsection{Experimental Setup}
\textbf{UtVAA} is instantiated in three configurations: \texttt{Tiny} (204.67K), \texttt{Medium} (609.36K), and \texttt{Large} (858.13K), as summarised in Table \ref{tab:modeldetails}. Experiments are conducted on \texttt{CIFAR} benchmark datasets and plant disease datasets to assess generalisation and practical performance. For deployment-oriented analysis, the \textit{Tiny} variant is prioritised due to its low computational complexity. All models are trained from scratch under identical settings to ensure a fair comparison with existing lightweight CNN and ViT-based architectures. Training is performed on a workstation with dual Intel\textsuperscript{\textregistered} Xeon\textsuperscript{\textregistered} 4215R CPUs and NVIDIA RTX A6000 GPUs (48 GB VRAM). To reflect real-world constraints, inference latency is also measured on an ARM-based Apple M1 device. Dataset-specific parameter tuning for \texttt{CIFAR} and plant disease datasets is detailed in the corresponding subsections.

\begin{table*}[h]
\centering
\caption{Performance comparison of \textbf{UtVAA} variants and existing lightweight models trained from scratch on CIFAR-10 and CIFAR-100 datasets.}

\label{tab:cifar}
\renewcommand{\arraystretch}{1.6}
\setlength{\tabcolsep}{4pt}
\footnotesize
\begin{tabular}{l l r r c c c c c}
\hline
\textbf{} & \textbf{Model} & \textbf{Params. (K)} & \textbf{FLOPs (M)} & 
\textbf{Top-1} & \textbf{Precision} & \textbf{Recall} & \textbf{F1-Score} & \textbf{Infer. (ms)} \\
\hline
\multirow{10}{*}{\rotatebox[origin=c]{90}{\textbf{CIFAR-10}}}
& MobileViT-V1\cite{mehta2021mobilevit} & 1,059.84 & 449.24 & 0.9219 & 0.9221 & 0.9219 & 0.9217 & 43.30 \\
& MobileNet-V2\cite{sandler2018mobilenetv2}& 2,236.68 & 399.95 & 0.8915	& 0.8914	& 0.8915 & 0.8909 & 44.39 \\
& \text{ShuffleNet-V2\_0.5} \cite{ma2018shufflenet} & 352.04 & \textbf{53.61} & 0.9096 & 0.9093 &	0.9096 & 0.9094 & 13.37 \\
& \text{SqueezeNet\_1.0}\cite{iandola2016squeezenet} & 740.56 & 973.55 & 0.8583 & 0.8581 & 0.8583 & 0.8575 & 33.86 \\
& DeiT\cite{touvron2021training} & 5,537.87 & 1,406.30 & 0.8872 & 0.8881 & 0.8872 & 0.8875 & 22.62 \\
& PiT\cite{heo2021rethinking} & 4,607.69 & 652.70 & 0.8783 & 0.8803 & 0.8783 & 0.8784 & 21.76 \\
& PSLT-Tiny \cite{wu2023pslt} & 3,917.84 & 874.40 & 0.9183 & 0.9184 & 0.9183 & 0.9177 & 93.14 \\
 \cline{2-9}
& UtVAA-Tiny & \textbf{178.66} & 53.92 & 0.9046 & 0.9042 & 0.9046 & 0.9042 & \textbf{10.07} \\
&UtVAA-Medium & 577.59 & 159.21 & 0.9264 & 0.9261 & 0.9264 & 0.9262 & 15.84 \\
& \textbf{UtVAA-Large} & 820.59 & 220.91 & \textbf{0.9334} & \textbf{0.9334} & \textbf{0.9334} & \textbf{0.9332} & 18.95 \\
\hline
\multirow{10}{*}{\rotatebox[origin=c]{90}{\textbf{CIFAR-100}}}
& MobileViT-V1\cite{mehta2021mobilevit} & 1,088.64 & 449.27 & 0.6904 & 0.6921 & 0.6904 & 0.6879 & 45.03 \\
& MobileNet-V2\cite{sandler2018mobilenetv2} & 2,351.97 & 400.07 & 0.6890 & 0.6881 & 0.6890 & 0.6858 & 50.80 \\
& \text{ShuffleNet-V2\_0.5} \cite{ma2018shufflenet} & 352.04 &	\textbf{53.61} & 0.6695	& 0.6801 &	0.6695 &	0.6669	& 14.06 \\
& \text{SqueezeNet\_1.0}\cite{iandola2016squeezenet} & 786.72 & 983.95 & 0.6141 & 0.6216 & 0.6141 & 0.6112 & 28.20 \\
& DeiT\cite{touvron2021training} & 5,555.24 & 1,406.30 & 0.6725 & 0.6865 & 0.6725 & 0.6693 & 32.25 \\
& PiT\cite{heo2021rethinking}& 4,630.82 & 652.70 & 0.6181 & 0.6260 & 0.6181 & 0.6110 & 38.43 \\
& PSLT-Tiny\cite{wu2023pslt}& 3,952.49 & 874.45 & 0.6959 & \textbf{0.7120} & 0.6959 & 0.6913 & 71.13 \\
 \cline{2-9}
& UtVAA-Tiny & \textbf{204.67} & 53.95 & 0.6459 & 0.6557 & 0.6459 & 0.6360 & \textbf{12.80} \\
& UtVAA-Medium & 609.36 & 159.24 & 0.6994 & 0.7070 & 0.6994 & 0.6973 & 22.83 \\
& \textbf{UtVAA-Large} & 858.13 & 220.94 & \textbf{0.7097} & 0.7100 & \textbf{0.7097} & \textbf{0.7066} & \textbf{27.33} \\
\hline

\end{tabular}
\end{table*}

\subsection{Results on CIFAR Benchmark Datasets}
The CIFAR-10 and CIFAR-100 datasets \cite{krizhevsky2009learning}, comprising 10 and 100 classes, are used to evaluate the generalisation capability of the proposed \textbf{UtVAA} model. The CIFAR-10 dataset includes 50,000 training images and 10,000 testing images, with the training set further split into 90\% for training and 10\% for validation. The same partitioning strategy is applied to CIFAR-100. Each image has an original resolution of $32 \times 32$ pixels. To improve generalisation, data augmentation techniques are applied, including resizing to $256 \times 256$ using InterpolationMode, followed by RandomCrop, RandomHorizontalFlip, RandomRotation, ColorJitter, RandAugment, RandomErasing, CutMix \cite{yun2019cutmix} and MixUp \cite{zhang2017mixup}. Training is conducted for up to 1000 epochs using AdamW with cosine learning rate decay. The learning rate is scheduled from $7 \times 10^{-4}$ to $1 \times 10^{-5}$ (or $1 \times 10^{-6}$ for CIFAR-10), with a weight decay of 0.01 and label smoothing of 0.1. Early stopping is applied to reduce overfitting.

\begin{figure}[h]
    \centering
    \includegraphics[width=\linewidth]{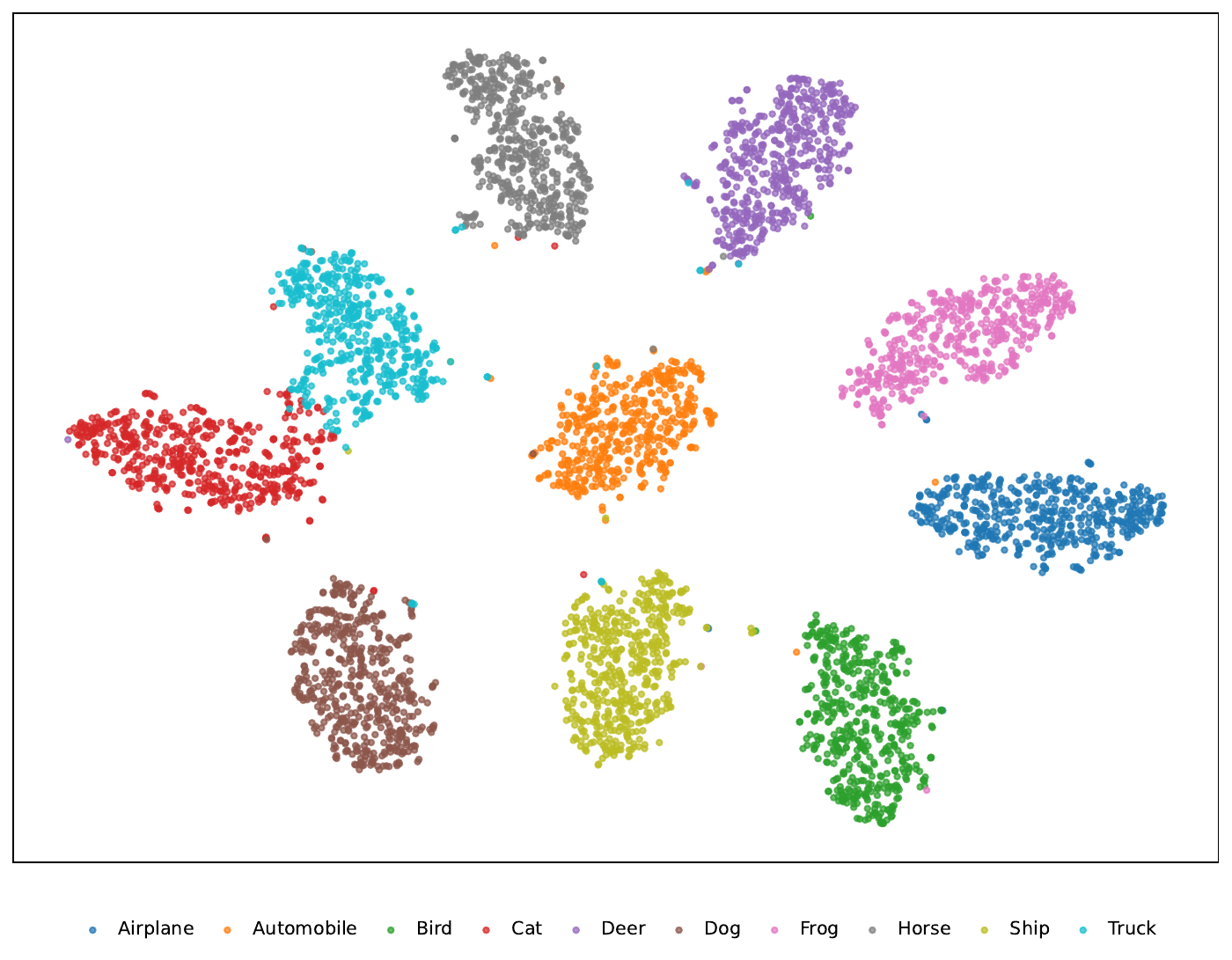}
    \caption{t-SNE visualisation of CIFAR-10 feature embeddings learned by the UtVAA architecture, showing clear class-wise clustering.}\label{fig:cifer10tsne}
\end{figure}

\begin{figure}
    \centering
    \includegraphics[width=\linewidth]{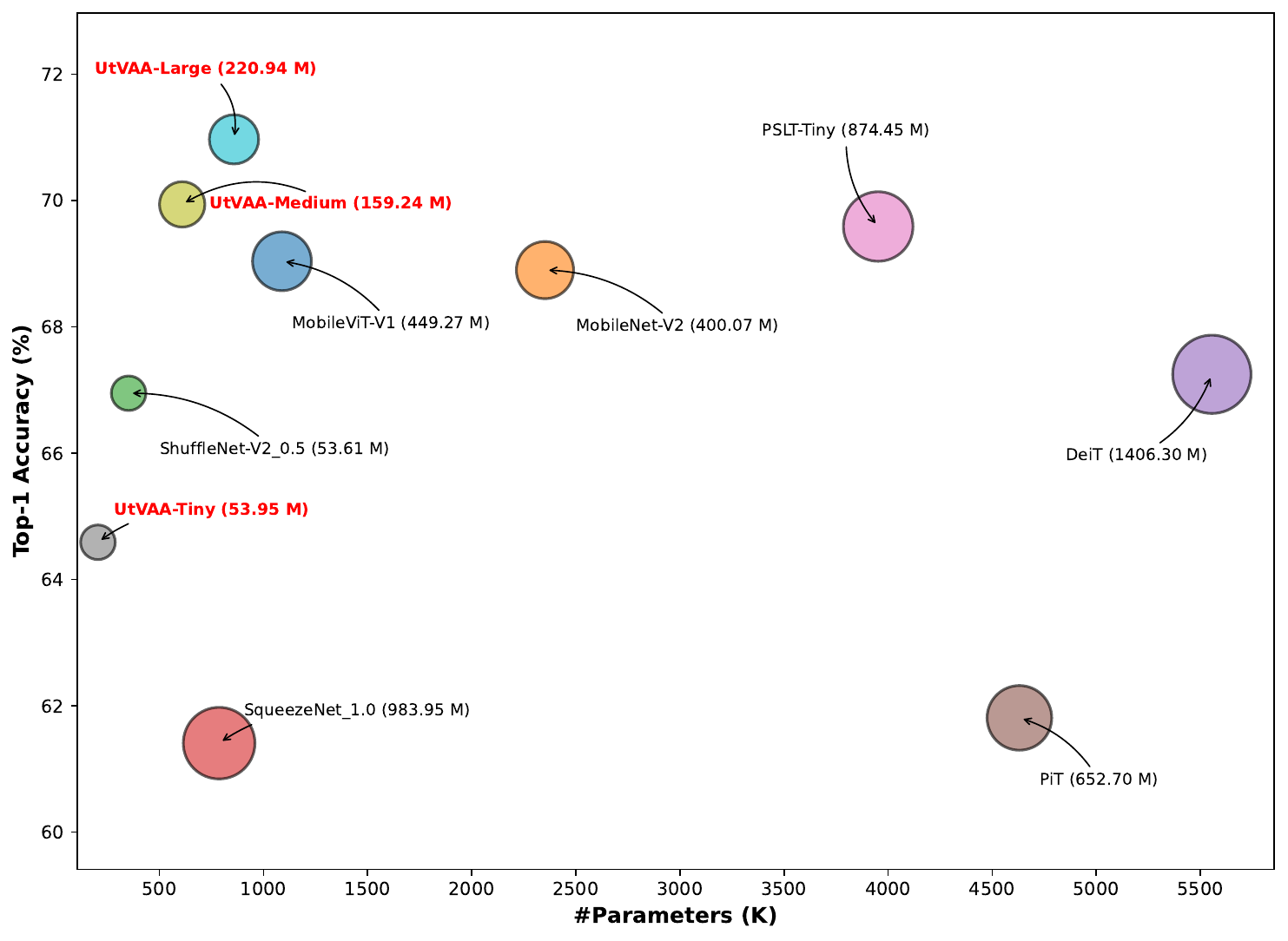}
    \caption{Top-1 accuracy comparison between UtVAA variants (Large, Medium, and Tiny) and existing lightweight CNN models on the CIFAR-100 dataset. The bubble size represents the number of FLOPs, as indicated after each model name.}
    \label{fig:parametervsaccuracy}
\end{figure}

Results on CIFAR-10 and CIFAR-100 for the proposed UtVAA architecture and existing state-of-the-art CNN- and ViT-based architectures are presented in Table \ref{tab:cifar}. The Top-1 accuracy results reinforce these observations. UtVAA-Large achieves the highest accuracy on both CIFAR-10 (0.9334) and CIFAR-100 (0.7097), outperforming all compared methods while maintaining a substantially lower parameter count than many transformer-based models. Fig. \ref{fig:cifer10tsne} presents the t-SNE visualisation for CIFAR-10, where UtVAA shows well-separated clusters with clear inter-class margins and compact intra-class distributions. Across both datasets, the UtVAA models achieve a balanced trade-off between model size and computational cost. In particular, UtVAA-Tiny has the lowest parameter count and FLOPs among all methods while maintaining competitive performance, which demonstrates its suitability under resource constraints. This trend is further reflected in Fig \ref{fig:parametervsaccuracy}, where the x-axis represents the number of parameters, the y-axis denotes Top-1 accuracy, and bubble size corresponds to FLOPs. The UtVAA variants occupy favourable regions in this space, combining higher accuracy with smaller model size and moderate computational cost. In particular, UtVAA-Large lies near the upper-left frontier compared with most baselines, indicating a more efficient accuracy-complexity trade-off.

As model capacity increases from Tiny to Medium and Large, performance improves consistently. A similar trend is observed for precision, recall and F1-score. On CIFAR-10, UtVAA-Medium and UtVAA-Large achieve higher and more balanced values across these metrics than most baseline models. The gap between precision and recall remains small, highlighting that the models do not bias specific classes. In one case, PSLT-Tiny achieved slightly higher precision. However, UtVAA outperformed it in all other metrics. The inference time highlights the efficiency of the proposed approach. UtVAA-Tiny achieves the lowest latency on both datasets, while UtVAA-Medium and UtVAA-Large maintain relatively low inference times despite increased capacity. Compared with models such as PSLT-Tiny and DeiT, which exhibit higher latency, the UtVAA variants offer a more suitable trade-off for real-time and resource-constrained applications. This efficiency is particularly relevant for deployment scenarios with computational and latency constraints.

Furthermore, the ablation study in Table \ref{tab:ablation} evaluates the contribution of the dilated bottleneck block, affix attention block and coordinate attention on the CIFAR-100 dataset. The UtVAA, incorporating the dilated bottleneck, affix attention block and coordinate attention, achieves the highest Top-1 accuracy of 0.7097. Removing the coordinate attention module reduces accuracy to 0.6791, with only a modest decrease in parameters and FLOPs, which reflects its contribution to spatial feature representation. A further-simplified variant that retains only the dilated bottleneck and excludes both attention mechanisms yields a larger decline in accuracy (0.6599), along with a significant reduction in model complexity. These results show that the dilated bottleneck serves as a strong baseline, while the inclusion of attention mechanisms, particularly coordinate attention, improves discriminative performance and supports the design of the UtVAA architecture.

\begin{table}[h]
\centering

\renewcommand{\arraystretch}{1.3}
\caption{Ablation study of UtVAA-Large with different architectural components.}
\label{tab:ablation}
\resizebox{0.5\textwidth}{!}{
\begin{tabular}{lcccrrr}
\hline
\textbf{Model} & \textbf{DB} & \textbf{Aff} & \textbf{CA} & \textbf{Params. (K)} & \textbf{FLOPs (M)} & \textbf{Top-1} \\ \hline
UtVAA-Large & \checkmark & \checkmark & \checkmark & 858.13 & 220.94 & 0.7097 \\
--          & \checkmark & \checkmark & \xmark     & 809.55 & 209.72 & 0.6791 \\
--          & \checkmark & \xmark     & \xmark     & 263.57 & 70.31  & 0.6599 \\ \hline
\end{tabular}
}

\vspace{5pt}
\begin{minipage}{0.5\textwidth} 
    \footnotesize
    \textbf{DB}: Dilated Bottleneck, 
    \textbf{Aff}: Affix Attention Block, 
    \textbf{CA}: Coordinate Attention
\end{minipage}
\end{table}

\subsection{Results on Plant Disease Datasets}
The field applicability of the UtVAA architecture is evaluated on two plant disease datasets, namely \texttt{PlantVillage}\footnote{https://github.com/spmohanty/plantvillage-dataset} and \texttt{SLIF-Tomato}\footnote{https://www.kaggle.com/datasets/romiyalgeorge/slif-tomato-dataset}. For the PlantVillage dataset, the tomato subset is used. For this analysis, UtVAA-Tiny is selected due to its low computational complexity among the UtVAA variants. The class distribution of the tomato subset of the PlantVillage dataset and the SLIF-Tomato dataset is presented in Table \ref{tab:pvslifdatasets}.

\begin{table}[h]
\centering
\caption{Class distribution of PlantVillage-Tomato and SLIF-Tomato datasets.}
\label{tab:pvslifdatasets}
\renewcommand{\arraystretch}{1.2}
\setlength{\tabcolsep}{6pt}
\begin{tabular}{lcc}
\toprule
\textbf{Class Name} & \textbf{PlantVillage-Tomato} & \textbf{SLIF-Tomato} \\

\midrule
Bacterial Spot & 2127 & 917 \\
Early Blight & 1000 & 1539 \\
Late Blight & 1909 & 954 \\
Mosaic Virus & 373 & 1544 \\
Septoria Leaf Spot & 1771 & 698 \\
Healthy & 1591 & 1243 \\
Target Spot & 1404 & \textemdash \\
Yellow Leaf Curl Virus & 5357 & \textemdash \\
Leaf Mold & 952 & \textemdash \\
Spider Mites Two Spotted & 1676 & \textemdash \\
Wilt & \textemdash & 1028 \\
Powdery Mildew & \textemdash & 1011 \\
\midrule
\textbf{Total} & \textbf{18160} & \textbf{8934} \\
\bottomrule
\end{tabular}
\end{table}

On the lab-based PlantVillage-Tomato dataset, which includes ten classes covering multiple disease types and healthy leaves, the model is trained and evaluated using 12,712 training images (70\%), 3,632 validation images (20\%), and 1,816 testing images (10\%). To assess performance in practical settings, the model is also evaluated on the SLIF-Tomato in-field dataset, which comprises eight classes with both diseased and healthy categories, using 6,253 training images (70\%), 1,786 validation images (20\%), and 895 testing images (10\%). For both datasets, images are resized to $256 \times 256$ pixels and augmented using RandomHorizontalFlip and RandomRotation to improve generalisation. Training is conducted for 400 epochs using the AdamW optimiser with a cosine learning rate decay schedule, where the learning rate ranges from $1 \times 10^{-3}$ to $1 \times 10^{-5}$. A weight decay of 0.001 and label smoothing of 0.01 are applied to stabilise training, and early stopping is used to reduce overfitting.

\begin{table*}[h]
\centering
\footnotesize
\caption{Performance comparison of UtVAA-Tiny and existing lightweight models trained from scratch on PlantVillage-Tomato and SLIF-Tomato datasets.}
\renewcommand{\arraystretch}{1.3}
\label{tab:PVSLIF}

\begin{tabular}{llrrcccccc}
\toprule
\textbf{} &\textbf{Model} & \textbf{Params. (K)} & \textbf{FLOPs (M)}  & \textbf{Top-1} & \textbf{Precision} & \textbf{Recall} & \textbf{F1-Score} & \textbf{Infer. (ms)} \\
\midrule

\multirow{8}{*}{\rotatebox[origin=c]{90}{\textbf{PlantVillage-Tomato}}}
& MobileViT-V1     & 1,059.84 & 449.24  & 0.9928 & 0.9912 & 0.9915 & 0.9913 & 37.57 \\
& MobileNet-V2     & 2,236.68 & 399.95  & 0.9945 & 0.9926 & 0.9929 & 0.9927 & 41.42 \\
& ShuffleNet-V2\_0.5 & 352.04 & \textbf{53.61}  & 0.9950 & 0.9896 & 0.9937 & 0.9915 & 12.86 \\
& SqueezeNet\_1.0  & 740.55 & 973.55 & 0.9901 & 0.9866 & 0.9874 & 0.9869 & 36.41 \\
& DeiT             & 5,537.87 & 1406.30 & 0.9774 & 0.9722 & 0.9736 & 0.9727 & 14.97 \\
& PiT              & 4,607.69 & 652.70  & 0.9813 & 0.9805 & 0.9789 & 0.9797 & 11.53 \\
& PSLT-Tiny        & 3,917.84 & 874.40  & 0.9939 & 0.9941 & 0.9926 & 0.9933 & 71.79 \\
\cline{2-9}
& \textbf{UtVAA-Tiny} & \textbf{178.66} & 53.92  & \textbf{0.9961} & \textbf{0.9957} & \textbf{0.9955} & \textbf{0.9956} & \textbf{8.41} \\

\midrule

\multirow{8}{*}{\rotatebox[origin=c]{90}{\textbf{SLIF}}}
& MobileViT-V1     & 1,059.20 & 449.24  & 0.9944 & 0.9943 & 0.9958 & 0.9950 & 37.69 \\
& MobileNet-V2     & 2,234.12 & 399.95 & 0.9922 & 0.9937 & 0.9948 & 0.9942 & 42.25 \\
& ShuffleNet-V2\_0.5 & 349.99 & \textbf{53.61}  & 0.9944 & 0.9954 & 0.9962 & 0.9958 & 10.53 \\
& SqueezeNet\_1.0  & 739.53 & 973.30  & 0.9832 & 0.9848 & 0.9854 & 0.9850 & 34.13 \\
& DeiT             & 5,537.48 & 1406.30  & 0.9832 & 0.9841 & 0.9880 & 0.9860 & 11.47 \\
& PiT              & 4,607.18 & 652.70  & 0.9844 & 0.9852 & 0.9881 & 0.9866 & 12.23 \\
& PSLT-Tiny        & 3,917.07 & 874.40 & 0.9821 & 0.9845 & 0.9829 & 0.9835 & 75.70 \\
\cline{2-9}
& \textbf{UtVAA-Tiny} & \textbf{178.09} & 53.92  & \textbf{0.9966} & \textbf{0.9973} & \textbf{0.9976} & \textbf{0.9974} & \textbf{7.45} \\
\bottomrule
\end{tabular}
\end{table*}

\begin{figure}[h]
    \includegraphics[width=\linewidth]{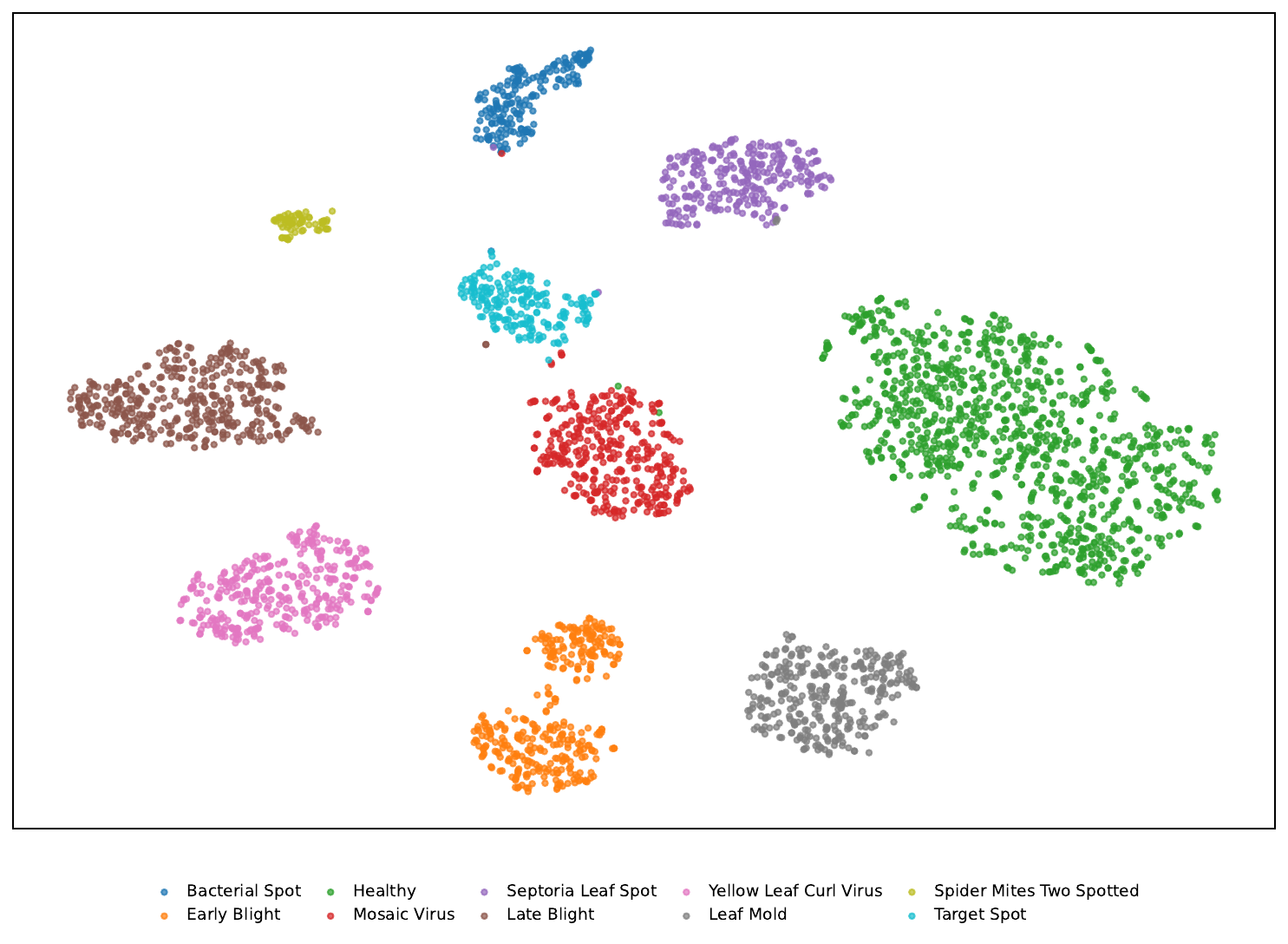}
    \caption{t-SNE visualisation for the PlantVillage-tomato dataset by UtVAA architecture.}\label{fig:PVtsne}
\end{figure}

\begin{figure}[h]
    \includegraphics[width=\linewidth]{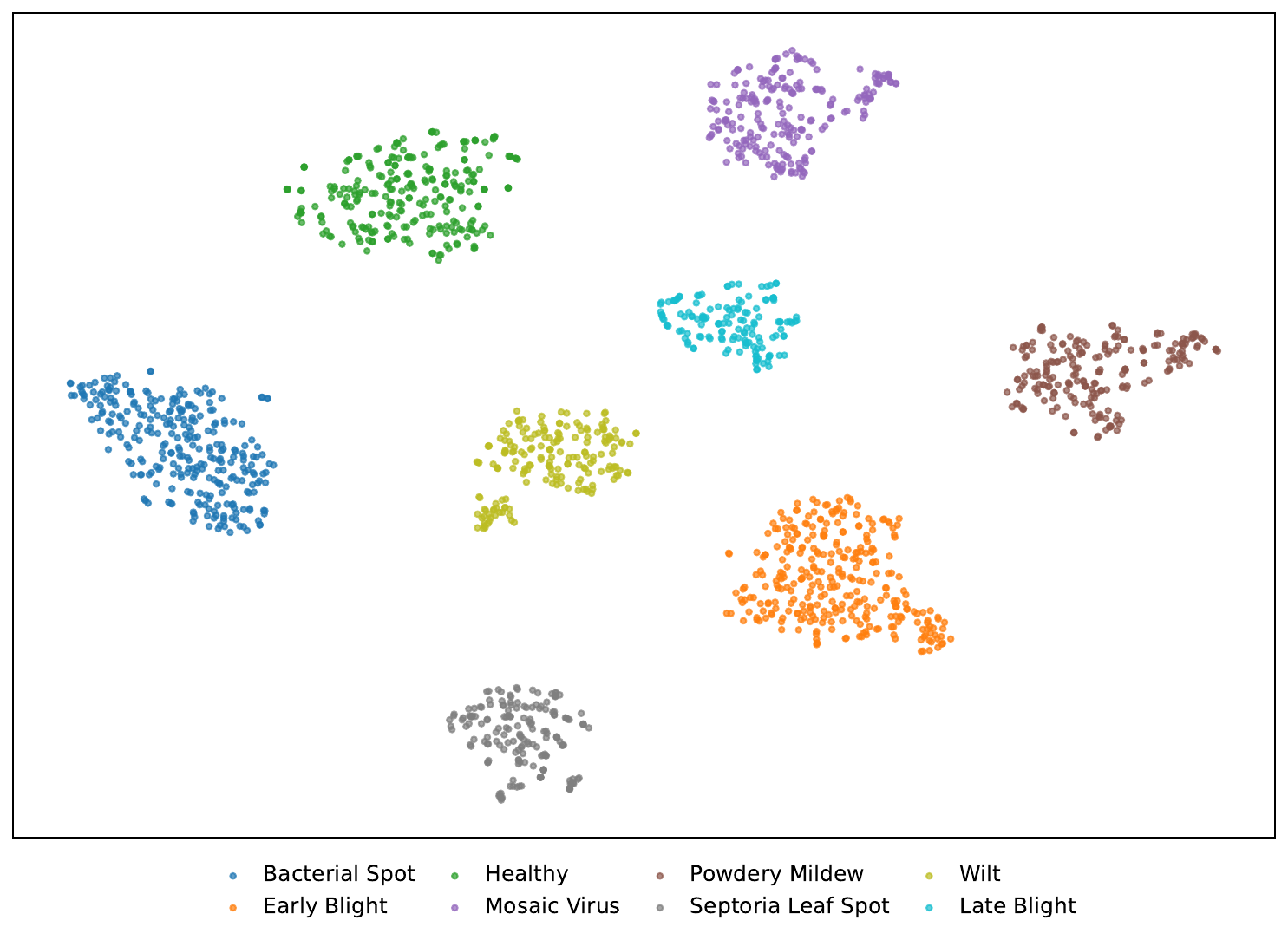}
    \caption{t-SNE visualisation for the SLIF-tomato dataset by UtVAA architecture.}\label{fig:SLIFtsne}
\end{figure}

The results in Table \ref{tab:PVSLIF} show that UtVAA-Tiny achieves a strong balance between predictive performance and computational efficiency across both datasets. On PlantVillage-Tomato, the model attains the highest Top-1 accuracy (0.9961) along with consistently high precision, recall, and F1-score, while using the fewest parameters (178.66K) among all compared methods. A similar pattern is observed on the SLIF-Tomato dataset, where UtVAA-Tiny again surpasses all baselines, achieving a Top-1 accuracy of 0.9966 and strong classification metrics. The accuracy of the UtVAA-Tiny model on both datasets is further supported by the t-SNE plots shown in Fig. \ref{fig:PVtsne} for PlantVillage-Tomato and Fig. \ref{fig:SLIFtsne} for the SLIF-Tomato dataset. These results are obtained with computational costs comparable to the most efficient baseline, ShuffleNet-V2\_0.5, but with higher predictive performance. This reflects the ability of the proposed architecture to capture more discriminative features under strict parameter and FLOPs constraints.

However, these results should be interpreted in the context of dataset characteristics. The consistently high performance across most models, particularly on PlantVillage-Tomato, suggests a less challenging task due to controlled imaging conditions and limited variability. As a result, performance differences are small, with only incremental improvements across models. In contrast, the SLIF-Tomato dataset, which reflects in-field conditions, offers a more realistic evaluation scenario. UtVAA-Tiny maintains superior performance under these conditions, which demonstrates stronger generalisation across varying backgrounds and lighting. In addition, the lower inference time of UtVAA-Tiny compared with transformer-based models such as DeiT, PiT and PSLT-Tiny supports its suitability for real-time agricultural applications, where both accuracy and latency are important.

\begin{figure}[ht]
    \centering
    \includegraphics[width=\linewidth]{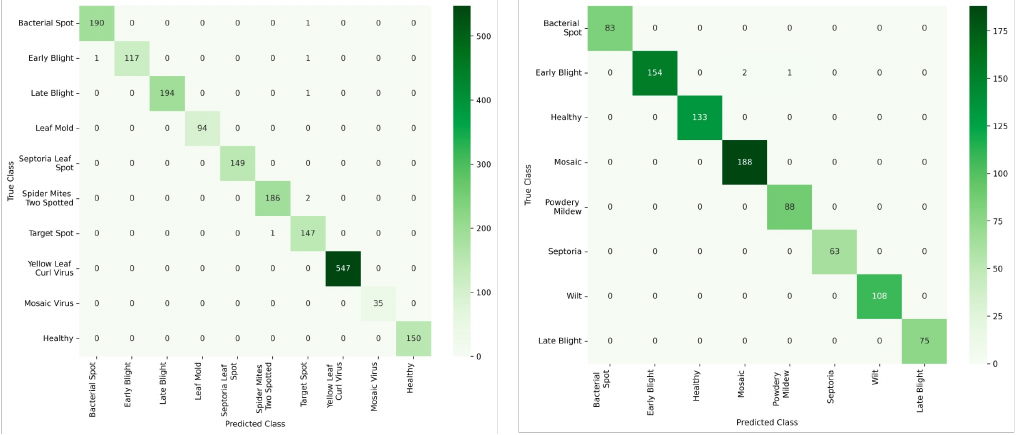}
    \caption{Confusion matrix for (left) PlantVillage-Tomato dataset and (right) SLIF-Tomato dataset.}
    \label{fig:confmatrix}
\end{figure}

Furthermore, the confusion matrices in Fig. \ref{fig:confmatrix} illustrate the class-wise performance of the UtVAA architecture on both PlantVillage-Tomato (left) and SLIF-Tomato (right). On the PlantVillage-Tomato dataset, the model achieves near-perfect classification for most classes, including Yellow Leaf Curl Virus, Leaf Mold and Healthy, with limited confusion observed between visually similar diseases such as Spider Mites Two-Spotted and Target Spot, and between Bacterial Spot and Target Spot. On the SLIF-Tomato dataset, similarly strong diagonal dominance is observed, with perfect classification for most classes, including Mosaic, Healthy, Powdery Mildew, Wilt, and Late Blight. Minor misclassifications occur only for Early Blight, which is occasionally predicted as Mosaic and Powdery Mildew. Overall, the results indicate strong discriminative capability, although class imbalance-particularly in the PlantVillage-Tomato dataset may influence the reported overall accuracy.

\begin{figure}
    \centering
    \includegraphics[width=\linewidth]{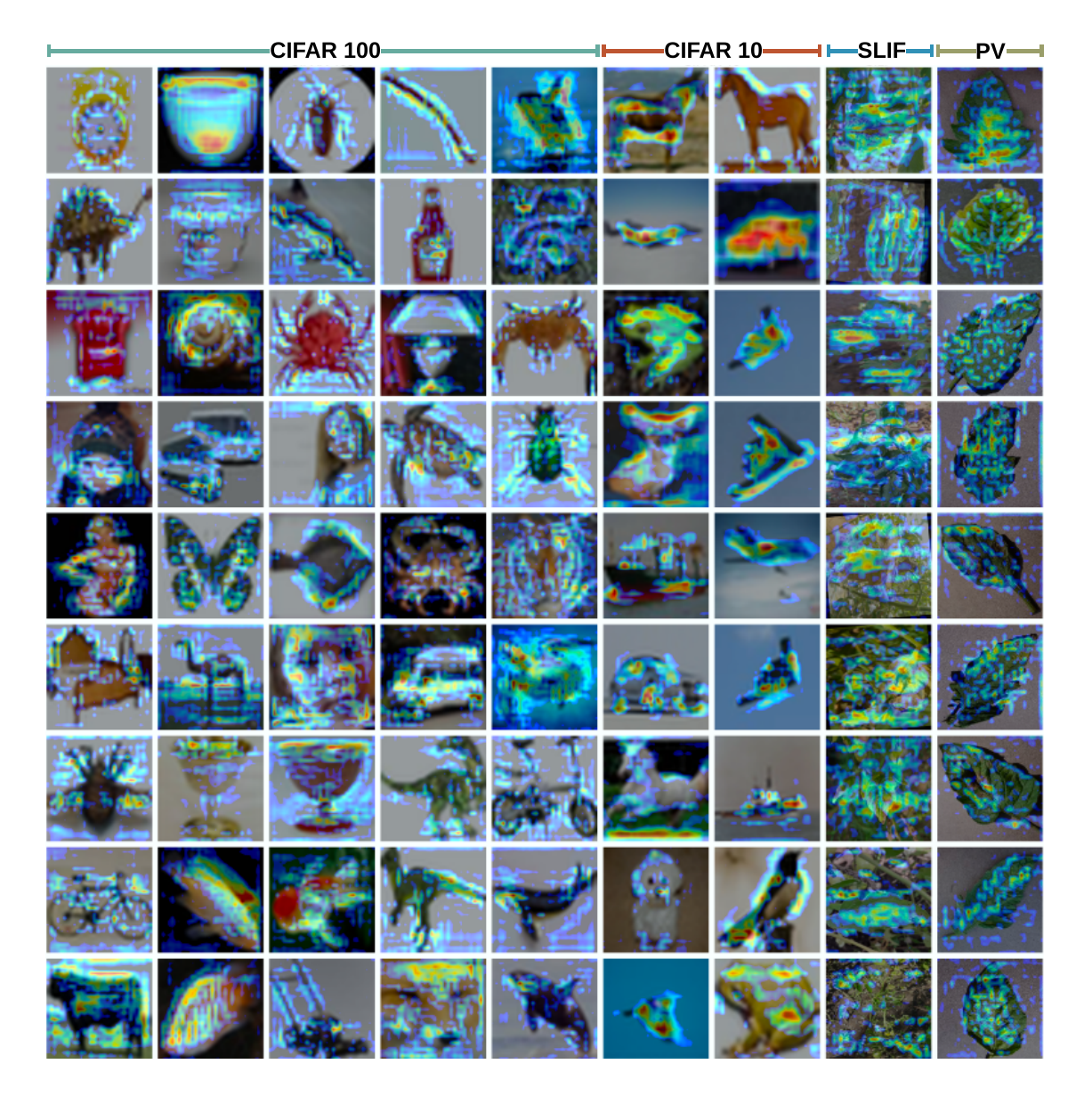}
    \caption{Grad-CAM visualisation: From left to right: CIFAR-100, CIFAR-10, SLIF-Tomato and PlantVillage-Tomato}
    \label{fig:grad_cam}
\end{figure}

Fig. \ref{fig:grad_cam} presents Grad-CAM heatmaps that demonstrate the interpretability of the proposed approach. Evaluations on CIFAR-10 and CIFAR-100 confirm the model’s general feature extraction capability. This initial validation provides a basis for robust performance prior to fine-grained symptom recognition in precision agriculture using the SLIF-tomato and PlantVillage-tomato datasets.

\section{Conclusion}\label{sec:conclusion}
This paper presented UtVAA, an ultra-tiny Vision Transformer designed for efficient mobile image classification. It introduces the Affix Attention block with a coordinated attention mechanism and linear transformer formulation, combined with an efficient ternary fusion strategy to reduce the computational cost of self-attention. In addition, the proposed Dilated Bottleneck block expands the receptive field using depthwise separable convolutions and residual connections while maintaining very low complexity. Unlike post-hoc compression techniques that may reduce representational capacity, UtVAA achieves efficiency through a principled architectural redesign that balances accuracy and computational cost. The model operates in a sub-million-parameter regime, with the Tiny variant containing only 204.67K parameters and 53.95M FLOPs, making it suitable for edge and mobile deployment. Experimental results on CIFAR-10, CIFAR-100, PlantVillage-Tomato and SLIF-Tomato demonstrate that UtVAA achieves competitive performance compared to existing lightweight architectures while significantly reducing computational overhead. Overall, this work shows that Vision Transformers can be redesigned into ultra-compact forms without substantial loss in classification performance. The proposed Affix Attention mechanism offers an efficient way to incorporate global context modelling into lightweight networks, narrowing the gap between high-performing transformer models and real-world mobile deployment constraints. Future work will focus on hardware-aware optimisation to further improve real-world efficiency, including deployment and evaluation on microcontroller-class and other highly resource-constrained devices. It will also explore extending the proposed architecture to more complex dense prediction tasks such as object detection and semantic segmentation, with the aim of broadening the applicability of ultra-compact vision transformers in practical computer vision systems.
 
\bibliographystyle{IEEEtran}
\bibliography{refs}

\begin{IEEEbiography}[{\includegraphics[width=1in,height=1.25in,clip,keepaspectratio]{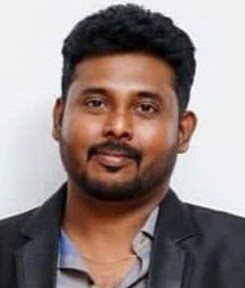}}]{Romiyal George}
 received M.Sc. degree from the University of Peradeniya and a B.Sc. degree from the University of Jaffna, Sri Lanka. He is currently a Ph.D. student at the University of Peradeniya, Sri Lanka.  He has worked in both industry and academia, and his research focuses on machine learning, deep learning, computer vision, and their applications in agriculture, emotion recognition, and steganography.
\end{IEEEbiography}

\begin{IEEEbiography}[{\includegraphics[width=1in,height=1.25in,clip,keepaspectratio]{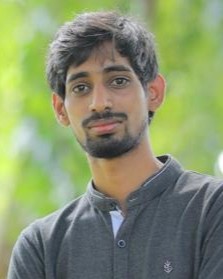}}]{Sathiyamohan Nishankar}
 received the BSc degree in Computer Engineering from the University of Peradeniya, Sri Lanka, in 2023. He is currently a Lecturer with the Faculty of Computing, Sabaragamuwa University of Sri Lanka. His research interests include machine learning, deep learning and computer vision, with applications including 3-dimensional image processing, medical imaging and explainable artificial intelligence.
\end{IEEEbiography}

\begin{IEEEbiography}[{\includegraphics[width=1in,height=1.25in,clip,keepaspectratio]{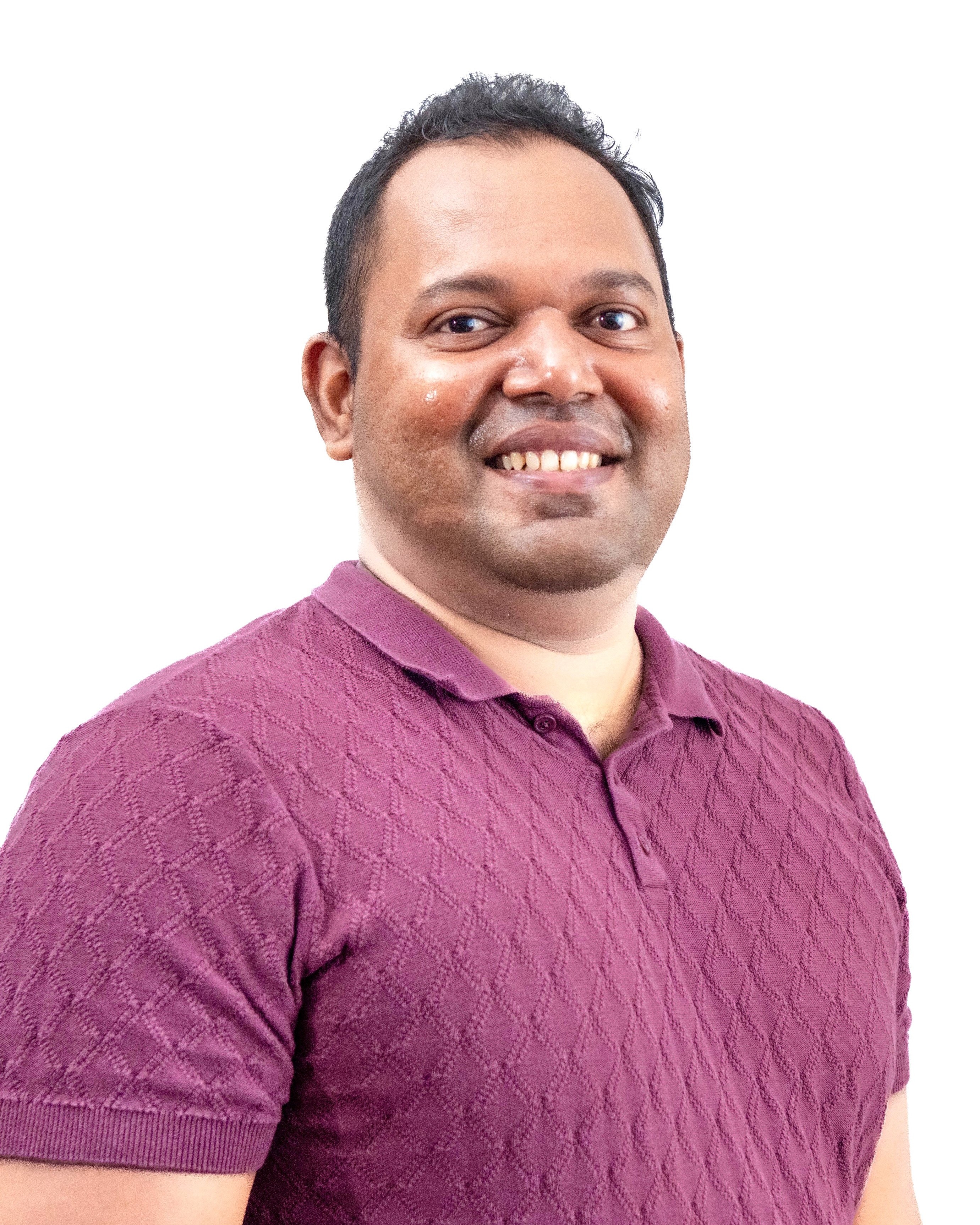}}]{Selvarajah Thuseethan}
	received the Ph.D. degree in information technology from Deakin University, Geelong, VIC, Australia, in 2022. He is currently a Lecturer with the Faculty of Science and Technology, Charles Darwin University, Casuarina, NT, Australia. He was previously a Postdoctoral Research Fellow with the School of Information Technology, Deakin University. His research interests include machine learning, deep learning, computer vision and their applications.
\end{IEEEbiography}

\begin{IEEEbiography}[{\includegraphics[width=1in,height=1.25in,clip,keepaspectratio]{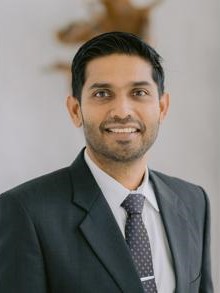}}]{Roshan G. Ragel}
	received his Ph.D. in Computer Science and engineering from UNSW Sydney, Sydney, NSW, Australia. He is a Professor with the Department of Computer Engineering, University of Peradeniya, Sri Lanka. His current research interests include systems-on-chip, the Internet of Things, accelerated and high-performance computing, computational biology, and wearable computing. He has been a Professional Member of the IEEE and IEEE Computer Society since 2005 and a Senior Member since 2014.
\end{IEEEbiography}

\end{document}